\pdfoutput=1

\documentclass[11pt]{article}

\usepackage[]{acl}

\usepackage{multicol}

\usepackage{times}
\usepackage{latexsym}
\usepackage{tcolorbox}

\usepackage[T1]{fontenc}

\usepackage{array}
\newcolumntype{P}[1]{>{\centering\arraybackslash}p{#1}}

\usepackage[utf8]{inputenc}

\usepackage{microtype}

\usepackage{bbm}
\usepackage{amsmath}
\usepackage{amssymb}

\usepackage{amsmath,amssymb,amsfonts}
\usepackage{bm}

\usepackage{xcolor,colortbl}
\usepackage{multirow}
\usepackage{booktabs}

\usepackage{graphicx}
\usepackage{subcaption}

\newcolumntype{P}[1]{>{\centering\arraybackslash}p{#1}}

\title{World Knowledge in Multiple Choice Reading Comprehension}

\author{Adian Liusie\thanks{\; Equal Contribution} \\
  ALTA Institute, Cambridge University \\
  \texttt{al826@cam.ac.uk} \\\And
Vatsal Raina\footnotemark[1] \\
  ALTA Institute, Cambridge University \\
  \texttt{vr311@cam.ac.uk} \\\AND
  Mark Gales \\
  ALTA Institute, Cambridge University \\
  \texttt{mjfg@cam.ac.uk} \\}

\begin{document}
 \maketitle
\begin{abstract}

Recently it has been shown that without any access to the contextual passage, multiple choice reading comprehension (MCRC) systems are able to answer questions significantly better than random on average. These systems use their accumulated "world knowledge" to directly answer questions, rather than using information from the passage. This paper examines the possibility of exploiting this observation  as a tool for test designers to ensure that the form of "world knowledge" is acceptable for a particular set of questions. We propose information-theory based metrics that enable the level of "world knowledge" exploited by systems to be assessed. Two metrics are described: the expected number of options, which measures whether a passage-free system can identify the answer to a question using world knowledge; and the contextual mutual information, which measures the importance of context for a given question. We demonstrate that questions with low expected number of options, and hence answerable by the shortcut system, are often similarly answerable by humans without context. This highlights that the general knowledge `shortcuts' could be equally used by exam candidates, and that our proposed metrics may be helpful for future test designers to monitor the quality of questions.

\end{abstract}

\section{Introduction}
Reading comprehension (RC) exams are used extensively in a wide range of language competency examinations \citep{AldersonJ.Charles.2000AR}, and have become a ubiquitous assessment method to probe how well candidates can read a passage and understand the text's core meaning. A fundamental assumption of RC exams is that to answer any of the questions correctly, one has to read the passage, comprehend its meaning, and identify the relevant information of a given question. However, recent work has shown that multiple-choice machine reading comprehension (MCMRC) systems without access to the passage can achieve reasonable performance \citep{pang-etal-2022-quality}, showing that the models may be doing less comprehension than anticipated.

\begin{figure}[t!]
    \centering
    \includegraphics[width=1.0\columnwidth]{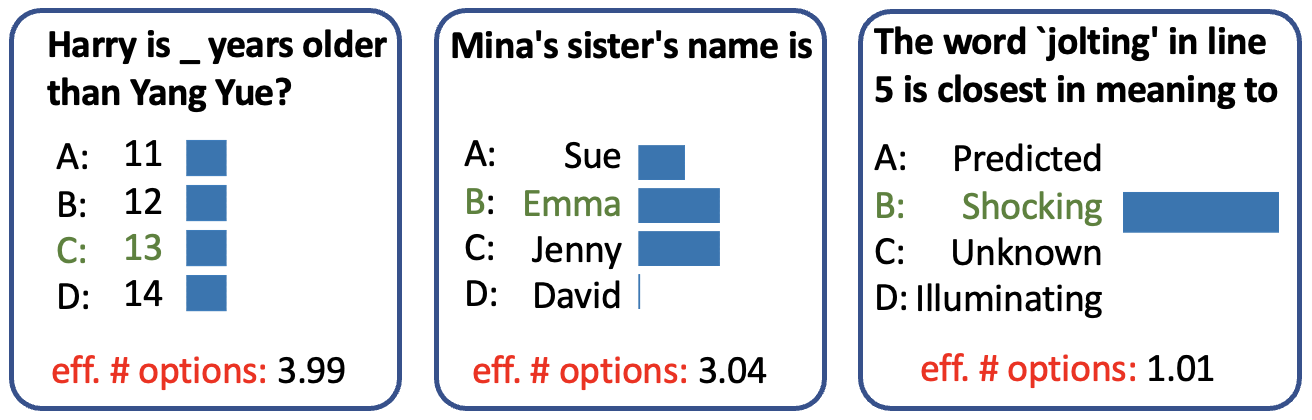}
    \caption{Output probabilities of our model (trained with contexts omitted) on real RACE++ \citep{pmlr-v101-liang19a} examples. `Effective number of options' is a metric that captures the model's confidence.}
    \label{fig:effective_options}
\end{figure}

\begin{figure*}[t!]
    \centering
    \includegraphics[width=\linewidth]{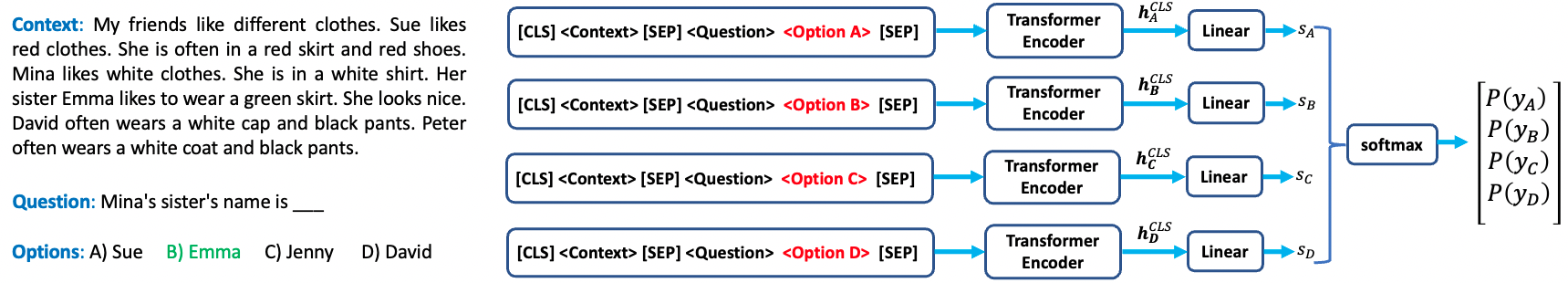}
    \caption{Model architecture.}
    \label{fig:model_arch}
\end{figure*}

In this paper we analyse this phenomena and for several standard MCMRC datasets \citep{pmlr-v101-liang19a, Huang2019CosmosQM, Yu2020ReClorAR} verify that passage-free baselines are able to achieve performance significantly better than random. We show that a subset of questions can be answered accurately and confidently without access to the contextual passage, where further analysis shows this is partly due to the presence of low-quality distractors, i.e. options that can be eliminated using only the question. As an example, given the question ``Mina's sister's name is:", one can eliminate any options that use a traditionally male name (see Figure \ref{fig:effective_options}). This highlights a potential `shortcut' candidates could use to answer questions while bypassing the context. Our work raises awareness to this potential flaw, and proposes a simple solution to catch questions that can be answered without comprehension. The proposed metrics might be a useful tool for future multiple-choice RC test designers to ensure that all questions truly assess reading comprehension ability.

Machine reading comprehension (MRC) is a highly researched area, with state-of-the-art (SoTA) systems \citep{Zhang2021RetrospectiveRF, Yamada2020LUKEDC, Zaheer2020BigBT, wang2021logicdriven} often approaching or even exceeding human level performance on public benchmarking leaderboards \citep{Clark2018ThinkYH, Lai2017RACELR, Trischler2017NewsQAAM, Yang2018HotpotQAAD}. Existing work has analysed the robustness of MRC systems, where researchers have questioned whether systems fully leverage context and whether they accomplish the underlying comprehension task \citep{sugawara2020assessing, rajpurkar-etal-2016-squad, kaushik2018much, jia2017adversarial, si2019does}. Most notably \citet{kaushik2018much} show that for a range of question-answering tasks, passage-only systems can often achieve remarkable performance, which has been observed in the MCRC setting \citep{pang-etal-2022-quality}.


Most existing work has discussed model robustness, demonstrating that for some tasks it is possible to obtain high average system performance with no context information. In contrast, this paper focuses on the attributes of individual questions and options, identifying questions where "world knowledge" can be leveraged, and the extent to which this knowledge can be leveraged. This could be a useful tool to enable test designers to monitor the questions being proposed, and whether alternative distractors or questions should be considered. 

\section{Multiple choice reading comprehension}
\label{sec:setup}
Multiple-choice reading comprehension is a popular task where given a context passage $C$ and question $Q$, the correct answer must be deduced from a set of answer options $\{O\}$. Current SoTA MRC systems are dominated by pre-trained language models (PrLMs) based on the transformer encoder architecture \citep{devlin2018bert, liu2020roberta, clark2020electra}. \\

\noindent\textbf{Model Architecture} Our question-answering system follows the standard MCMRC architecture of Figure \ref{fig:model_arch} \citep{Yu2020ReClorAR, raina-gales-2022-answer}.
Each option is individually encoded along with the question and the context into a score, and a softmax layer converts the 4 options' scores into a probability distribution. At inference, the predicted answer is the option with the greatest probability. \\

\begin{figure}[h!]
    \centering
    \includegraphics[width=\columnwidth]{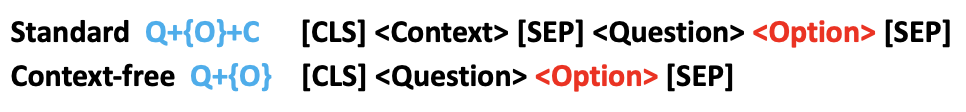}
    \caption{System inputs for shortcut system.}
    \label{fig:shortcut_inputs}
\end{figure}

\noindent\textbf{`No Context' Shortcut System} A requirement for good MCRC questions is that information from both the question and the context passage must be used to determine the correct answer. To probe whether this is an issue for MCMRC, we train systems using `context free' inputs (similar to \citet{pang-etal-2022-quality}). The standard set-up (Figure \ref{fig:model_arch}) is still followed, however the input is now altered to exclude the context, as shown in Figure \ref{fig:shortcut_inputs}.

\noindent\textbf{Effective Number of Options} Consider the output probability distribution of the predicted answer, ${\tt P}(y)$. One can determine the entropy, ${\cal H}(Y)$, which can be converted into the more interpretable \textit{effective number of options}, ${\cal N}(Y)$, a value bounded between 1 and the maximum number of options:
\begin{equation}
\label{eq:effective_options}
\small
\begin{aligned}
 {\cal N}(Y) = 2^{{\cal H}(Y)}, \hspace{5pt} {\cal H}(Y) = -\sum_{y\in Y} {\tt P}(y) \log_2 {\tt P}(y)
\end{aligned}
\end{equation}
For well designed questions, one would expect systems with missing information (i.e. the `shortcut' models) to have no information of what the answer is. This would correspond to a uniform distribution output (the distribution of maximum entropy), with an effective number of options equal to the total number of answer options. However, if the effective number of options is significantly lower than the total number of answer options, then this implies that prior information stored during training can be used to answer the question, without comprehension. \\

\noindent\textbf{Mutual Information} To probe how much information is gained by the context, one can additionally look at an approximation of mutual information of the context. This looks at how much the entropy decreases between the `no context' shortcut system and the `context' baseline system .
\begin{equation}
\label{eq:mutualinformation}
\small
\begin{aligned}
    \!\! {\cal I}(Y;C|Q, \{O\}) = 
    {\cal H}(Y|Q, \{O\}) - {\cal H}(Y|Q, \{O\}, C)
\end{aligned}
\end{equation}

\noindent An alternative approach would be to use random contexts \citep{creswell2022faithful} however we consider the stricter `no context' setting.

\section{Experiments}
\noindent\textbf{Data} We consider three popular MCMRC datasets: RACE++ \citep{Lai2017RACELR}, COSMOSQA \citep{Huang2019CosmosQM} and ReClor \citep{Yu2020ReClorAR}. RACE++ is a dataset of English comprehension questions for Chinese high school students, COSMOSQA is a large scale commonsense-based reading comprehension dataset, while ReClor is a challenging logical reasoning dataset at a graduate student level. All datasets have 4 options per question, one of which being the correct answer. \\

\begin{table}[htbp!]
\centering
\begin{small}
    \begin{tabular}{l|rrr}
    \toprule
& \texttt{TRN} & \texttt{DEV} & \texttt{EVL}  \\
\midrule
RACE++ & 100,388 & 5,599 & 5,642 \\
COSMOS & 25,262 & 2,985 & -- \\
ReClor & 4,638 & 500 & 1000 \\
   \bottomrule
    \end{tabular}
    \end{small}
    \caption{Dataset statistics}
    \label{tab:alldata}
\end{table}

\noindent\textbf{Training} An ELECTRA-large\footnote{\url{https://huggingface.co/docs/transformers/model_doc/electra}} model is fine-tuned on the training split \texttt{TRN}, hyper-parameters are tuned on the developement set \texttt{DEV}, and performance reported on the test split \texttt{EVL} for RACE++ (\texttt{DEV} splits are used for COSMOS and ReClor due to unavailability of the \texttt{EVL} splits). All model parameters (transformer and classifier) are updated during fine-tuning. Additionally, models are trained and evaluated using the `no context', as described in Section \ref{sec:setup}. Final hyperparameters are given in Appendix \ref{sec:train-det}. Three seeds are trained, and ensemble accuracy is used as the default metric when reporting performance.\footnote{code for experiments available at: \\ \url{https://github.com/adianliusie/MCRC}}

\subsection{Results}
\noindent\textbf{Context-Free Performance} We compare the performance of the baseline `standard' system against the shortcut `no context' systems for the various MCMRC datasets. Table \ref{tab:probe_cross} illustrates that the shortcut systems achieve high performance across all MCMRC datasets, all above 50\%, significantly above the expected random performance of 25\%. Further, we find that the shortcut rules can generalise across domains, most notably seen with the 54\% performance when training the shortcut system on RACE and evaluating on COSMOS. This highlights that the shortcut performance cannot be explained purely by dataset bias, but that there is a skill, unrelated to comprehension, that the systems are meaningfully leveraging. 

\begin{table}[ht]
\centering
\begin{small}
    \begin{tabular}{ll|ccc}
    \toprule
 \multicolumn{2}{c|}{Training data}  & RACE++ & COSMOS & ReClor \\
\midrule
& -- & 25.00 & 25.00 & 25.00 \\
\midrule
\multirow{2}{*}{RACE++}
& stan.   & \textbf{85.01} & 70.05 & 48.60 \\
& no con. & \textbf{57.32} & 54.04 & 34.80 \\
\midrule
\multirow{2}{*}{COSMOS}
& stan. & 66.81 & \textbf{84.49} & 41.20 \\
& no con.   & 38.73 & \textbf{68.51} & 27.80 \\
\midrule
\multirow{2}{*}{ReClor}
& stan. & 52.69 & 41.68 & \textbf{69.80} \\
& no con.   & 31.27 & 33.13 & \textbf{51.80} \\
   \bottomrule
    \end{tabular}
    \end{small}
\caption{Cross-performance of systems on RACE++, COSMOSQA and ReClor (standard vs no context).}
\label{tab:probe_cross}
\end{table}

\begin{figure}[h!]
    \centering
    \includegraphics[width=1.0\columnwidth]{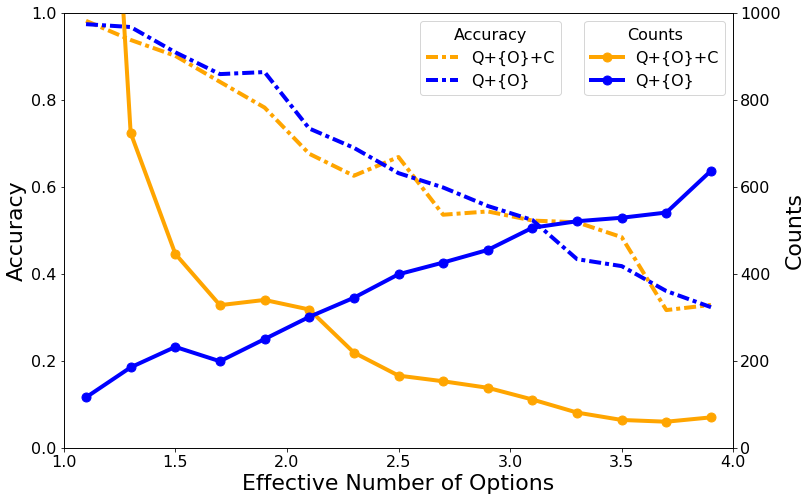}
    \caption{Distribution of effective number of options and corresponding (binned) accuracy.}
    \label{fig:counts_acc_plot}
\end{figure}

\noindent\textbf{RACE++ Effective Number of Options} Figure \ref{fig:counts_acc_plot} presents the count and accuracy plots of the effective number of options (bin width of 0.2) for both the standard and shortcut systems on RACE++ (see Appendix for other datasets). The counts plot show the number of questions within the bin range, while accuracy refers to the accuracy over all the examples within the bin. Since the systems are slightly overconfident\footnote{For both models, the mean of the maximum probability is 5\% above the overall accuracy.}, the systems' output probabilities are calibrated using temperature annealing \citep{guo2017calibration} (see Appendix \ref{sec:calibration}). 

The baseline system has high certainty for most points, which is somewhat expected given the baseline's high accuracy. However the shortcut system, without any contextual information, has a significant number of examples in the very low entropy region. This shows that for a subset of questions, the system can confidently answer questions correctly without doing any comprehension at all. In other cases, the shortcut system can leverage some information from the question and can reduce the number of effective options to between 2-3, which implies that certain poor distractors can be eliminated by the question alone. We also show that for both models, there is a clear linear relationship between uncertainty and accuracy, illustrating that the context-free system's use of world knowledge is sensible and that it leverages meaningful task information (see \ref{sec:more_ex} for low-entropy examples). This confirms that the systems are well calibrated and that the effective number of options is a good measure of actual model uncertainty. \\

\begin{figure}[t]
    \centering
    \includegraphics[width=1.0\columnwidth]{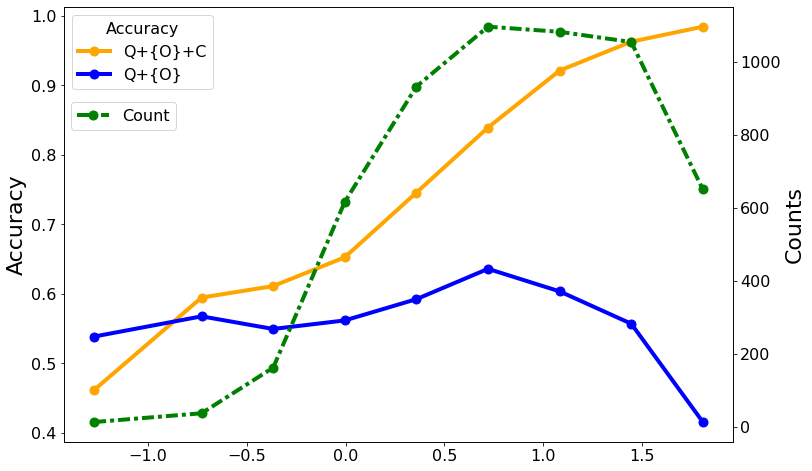}
    \caption{Distribution of counts and corresponding accuracy when points are sorted by MI approximation.}
    \label{fig:mut_info_plot}
\end{figure}

\noindent\textbf{Mutual Information} To further look at the influence of context, the mutual information (MI) between prediction and context was approximated for each example using Equation \ref{eq:mutualinformation}. Examples with a high MI are questions where the model is certain of the answer with context, but is uncertain without context - a desired property for comprehension questions.
Figure \ref{fig:mut_info_plot} shows the counts when all the examples are ordered by MI (see Equation \ref{eq:mutualinformation}) along with both the baseline and shortcut system accuracies. We note that the count distribution has a mix of high and low MI questions, which shows that the benefit of context is not a system-wide property but instead varies over questions. The accuracy of the baseline system increases considerably when context is useful, while accuracy falls for the shortcut system. It is interesting that a small fraction of questions have negative MI. Though MI should always be positive, negative values can be observed since models are only approximations of the true underlying distributions. The low accuracy of the shortcut model on negative MI questions occurs when standard world knowledge is not consistent with the information in the context. \\

\noindent\textbf{Human Evaluation of Metrics} 
We perform human evaluation to judge the practical use of our metrics.
 We select 100 questions with lowest and highest entropy, and three volunteer graduate students independently answer the questions without access to the context. We further select 50 questions with lowest and highest MI, and get our volunteers to first answer questions without context, then with context, and calculate the accuracy increase. All questions are shuffled, and volunteers attempt to best answer all questions. We find that our metrics are very effective in measuring their desired properties. Without context, humans are often able to answer the questions that the shortcut systems answer confidently, with humans achieving an average accuracy of 92\% on the 100 lowest entropy and 32\% on the 100 highest entropy examples respectively. Further, for high MI questions humans get a performance boost of 71\% when context is provided, and only 22\% for low mutual information questions.

\begin{table}[ht]
    \centering
    \begin{small}
    \begin{tabular}{l|cc|cc}
    \toprule
     & low ent. & high ent. & high MI & low MI \\
    \midrule
    human    & $91.7_{\pm1.9}$ & $31.7_{\pm2.9}$ & $\Delta 69.3_{\pm0.9}$ & $\Delta 24.7_{\pm5.0}$ \\
    system   & $99.0_{\pm0.0}$ & $24.3_{\pm6.2}$ & $\Delta 68.0_{\pm0.9}$ & $\Delta 3.3_{\pm 4.7}$ \\
    \bottomrule
    \end{tabular}
    \end{small}
\caption{Human and system `no context' accuracy on lowest and highest entropy questions as well as human and system change in accuracies on lowest and highest mutual information questions.}
\label{tab:human_eval}
\end{table}

\section{Conclusions}

For popular MCMRC datasets, systems can achieve reasonably high performance without performing any comprehension. Without passage information, `shortcut' systems can confidently determine some correct answer options, eliminate some unlikely distractors, and use general knowledge to gain information. 
Rather than focusing on average system performance, our work analyses individual question's reliance on world knowledge. We propose a metric based on the shortcut systems to automatically flag questions that are answerable without comprehension. We further provide evidence that the flagged questions are answerable by humans without any context. Lastly, using an approximation of the mutual information, we show that the importance of context varies over the questions in the dataset, and reason that high MI questions can be thought of as candidates for high-quality questions that truly measure comprehension abilities.

\section{Limitations}
We propose an approach that can automatically flag questions that can be answered without contextual information. However, the remaining questions are not necessarily high-quality questions, since many other aspects make up question quality. Second, the experiments are conducted using only the Electra model, though it is expected similar trends will be picked up by alternative transformer-based language models. Further, exams might be aimed at a level where a lack of specific knowledge may be assumed. Our work does not consider variable candidate knowledge levels, and our evaluation was only done by highly educated (we'd like to think) graduate students. Finally, we acknowledge that our human evaluation was limited in size and questions, however it is clearly demonstrated that for low `shortcut entropy' questions, comprehension is not necessarily required. 

\section{Acknowledgements}
This research is funded by the EPSRC (The Engineering and Physical Sciences Research Council) Doctoral Training Partnership (DTP) PhD studentship and supported by Cambridge Assessment, University of Cambridge and ALTA. 

\section{Ethics Statement}
There are no serious ethical concerns with this work. The human volunteers all performed the human evaluation tasks willingly without any coercion. The human evaluation took 2 hours per person.



\bibliography{anthology,custom}
\bibliographystyle{acl_natbib}

\appendix
\newpage

\renewcommand{\thesection}{\Alph{section}}
\renewcommand\thefigure{\thesection.\arabic{figure}} 
\setcounter{figure}{0}
\renewcommand\thetable{\thesection.\arabic{table}} 
\setcounter{table}{0}
\renewcommand{\thesection}{Appendix \Alph{section}}

\section{Additional Results} \label{sec:add-res}

\subsection{COSMOSQA}
\begin{figure}[h!]
    \centering
    \includegraphics[width=1.0\columnwidth]{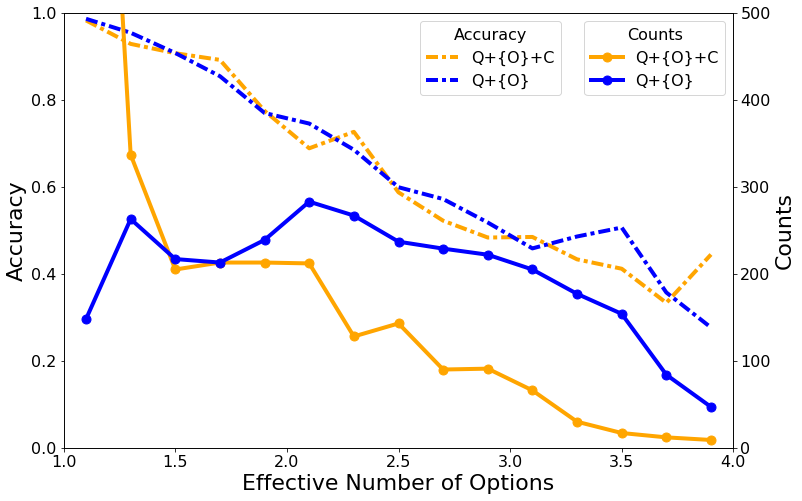}
    \caption{Distribution of effective number of options and binned accuracy for COSMOSQA.}
    \label{fig:cosmos_entropy}
\end{figure}

\begin{figure}[h!]
    \centering
    \includegraphics[width=1.0\columnwidth]{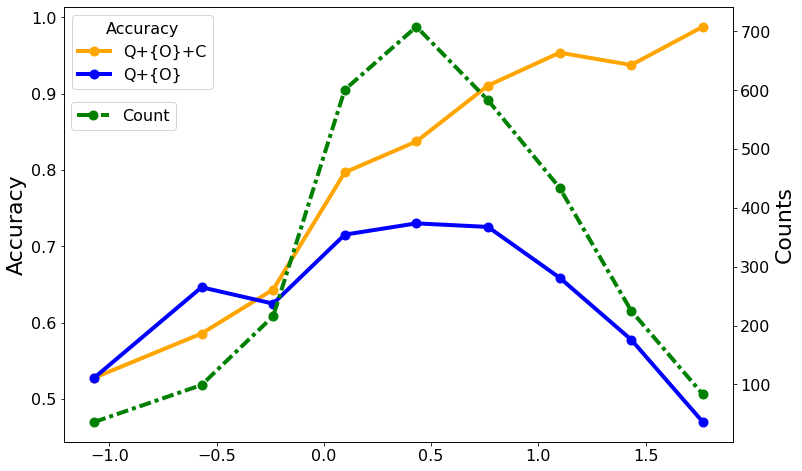}
    \caption{Distribution of counts and corresponding accuracy when points are sorted by MI approximation for COSMOSQA.}
    \label{fig:cosmos_mut_info}
\end{figure}

We repeat the entropy plot (Figure \ref{fig:cosmos_entropy}) for COSMOSQA and find similar trends to those seen in RACE++. The shortcut no-context system has a very flat distribution with a substantial number of questions answerable without context, with the effective number of options again having a clean linear relationship with accuracy. The repeated mutual information plot (Figure \ref{fig:cosmos_mut_info}) for COSMOSQA also has the same trend seen in RACE++, validating that our findings are more general that just for RACE++. 

\subsection{ReClor}
\begin{figure}[h!]
    \centering
    \includegraphics[width=1.0\columnwidth]{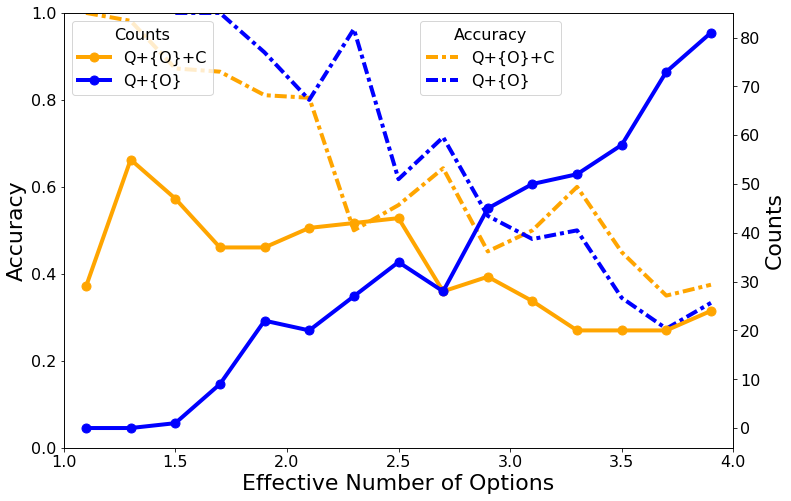}
    \caption{Distribution of effective number of options and binned accuracy for ReClor.}
    \label{fig:reclor_entropy}
\end{figure}

\begin{figure}[h!]
    \centering
    \includegraphics[width=1.0\columnwidth]{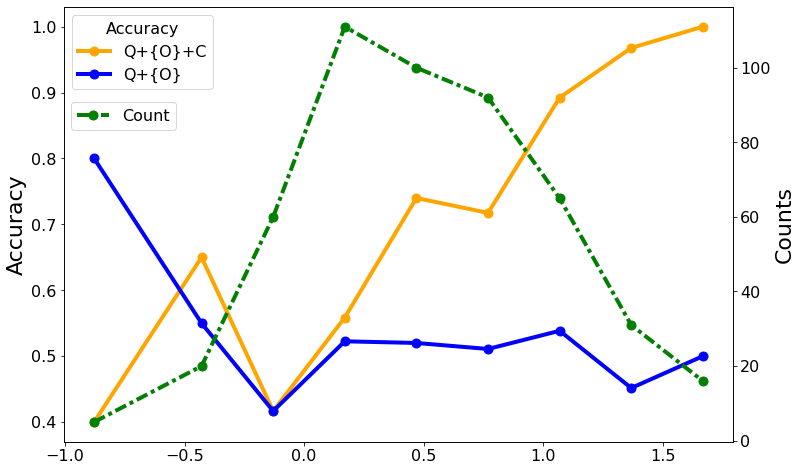}
    \caption{Distribution of counts and corresponding accuracy when points are sorted by MI approximation for ReClor.}
    \label{fig:reclor_mut_info}
\end{figure}

ReClor show roughly the same trends, however the questions of ReClor are much more challenging than in either RACE++ and COSMOSQA, and so we notice that the counts distribution is pushed considerably to the higher entropy side. Additionally, since ReClor is much smaller than RACE++ and COSMOSQA (see Table \ref{tab:alldata}), the curves are less smooth and largely suffer from noise. 

\subsection{Other Shortcuts}
\begin{figure}[h!]
    \centering
    \includegraphics[width=\columnwidth]{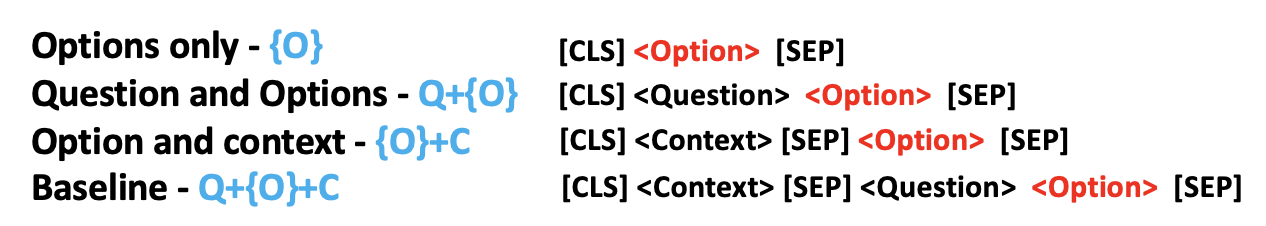}
    \caption{System inputs for alternative shortcut systems.}
    \label{fig:shortcut_inputs_all}
\end{figure}

We also consider other shortcut approaches, such as having context and options (i.e. missing question) and only options (Figure \ref{fig:shortcut_inputs_all}). Performance of the systems is shown in Table \ref{tab:cross_all_shortcut}. 

\begin{table}[ht]
\centering
\begin{small}
    \begin{tabular}{lc|ccc}
    \toprule
 \multicolumn{2}{c|}{Training data}  & RACE++ & COS. & ReClor \\
\midrule
& -- & 25.00 & 25.00 & 25.00 \\
\midrule
\multirow{4}{*}{RACE++}
& \{O\}     & \textbf{41.76} & 21.44 & 34.00 \\
& Q+\{O\}   & \textbf{57.32} & 54.04 & 34.80 \\
& \{O\}+C   & \textbf{68.20} & 54.61 & 46.00 \\
& Q+\{O\}+C & \textbf{85.01} & 70.05 & 48.60 \\
\midrule
\multirow{4}{*}{COSMOS}
& \{O\}     & 29.95 & \textbf{57.39} & 25.20 \\
& Q+\{O\}   & 38.73 & \textbf{68.51} & 27.80 \\
& \{O\}+C   & 52.41 & \textbf{78.96} & 40.40 \\
& Q+\{O\}+C & 66.81 & \textbf{84.49} & 41.20 \\
\midrule
\multirow{4}{*}{ReClor}
& \{O\}.    & 26.07 & 18.29 & \textbf{49.00} \\
& Q+\{O\}   & 31.27 & 33.13 & \textbf{51.80} \\
& \{O\}+C   & 39.83 & 36.88 & \textbf{68.40} \\
& Q+\{O\}+C & 52.69 & 41.68 & \textbf{69.80} \\
   \bottomrule
    \end{tabular}
    \end{small}
\caption{Cross-performance of systems on RACE++, COSMOSQA and ReClor using accuracy.}
\label{tab:cross_all_shortcut}
\end{table}

\newpage
\renewcommand{\thesection}{Appendix \Alph{section}}
\section{Model Information} \label{sec:all_details}
\renewcommand{\thesection}{\Alph{section}}
\renewcommand\thefigure{\thesection.\arabic{figure}} 
\setcounter{figure}{0}
\renewcommand\thetable{\thesection.\arabic{table}} 
\setcounter{table}{0}

\subsection{Training Details} \label{sec:train-det}
For all systems, deep ensembles of 3 models are trained with the  large~\footnote{Configuration at: \url{https://huggingface.co/google/electra-large-discriminator/blob/main/config.json}.} ELECTRA PrLM as a part of the multiple-choice MRC architecture depicted in Figure \ref{fig:model_arch}.
Each model has 340M parameters.
Grid search was performed for hyperparameter tuning with the initial setting of the hyperparameter values dictated by the baseline systems from \citet{Yu2020ReClorAR, raina-gales-2022-answer}. Apart from the default values used for various hyperparamters, the grid search was performed for the maximum number of epochs $\in \{2,5,10\}$; learning rate $\in \{2e-7, 2e-6, 2e-5\}$; batch size $\in \{2,4\}$.
For RACE++, training was performed for 2 epochs at a learning rate of 2e-6 with a batch size of 4 and inputs truncated to 512 tokens.
For systems trained on ReClor the final hyperparameter settings included training for 10 epochs at a learning rate of 2e-6 with a batch size of 4 and inputs truncated to 512 tokens.
For COSMOSQA, training was performed for 5 epochs at a learning rate of 2e-6 with a batch size of 4 and inputs truncated to 512 tokens.
Cross-entropy loss was used at training time with models built using NVIDIA A100 graphical processing units with training time under 3 hours per model for ReClor, 5 hours for COSMOSQA and 4 hours for RACE++.
All hyperparameter tuning was performed by training on TRN and selecting values that achieved optimal performance on DEV. For fairness, the `shortcut' systems (omitting various forms of the input) for each dataset were trained with the same hyperparameter settings as their corresponding baseline systems.

\subsection{Evaluation Details}
For each dataset, the systems are trained on the training split and hyperparameter tuned on the development split. For RACE++, systems are evaluated on the held out test data, but for COSMOSQA and ReClor, the evaluations are performed on the development split because their test splits have their labels hidden.

\subsection{Calibration}\label{sec:calibration}
The trained models were calibrated post-hoc using single parameter temperature annealing \citep{guo2017calibration}. Uncalibrated, model probabilities are determined by applying the softmax to the output logit scores $s_i$:
\begin{equation}
    P(y = k ; \bm{\theta}) \propto \exp(s_k)
\end{equation}
where $k$ denotes a possible output class for a prediction $y$.
Temperature annealing `softens' the output probability distribution by dividing all logits by a single parameter $T$ before the softmax. 
\begin{equation}
    P_{CAL}(y = k ; \bm{\theta}) \propto \exp(s_k/T)
\end{equation}
As the parameter $T$ does not change the relative rankings of the logits, the model's prediction will be unchanged and so temperature scaling does not affect the model’s accuracy. The parameter $T$ is chosen such that the accuracy of the system is equal to the mean of the maximum probability (which would be expected for a calibrated system).

\renewcommand{\thesection}{Appendix \Alph{section}}
\section{Licenses} \label{sec:licenses}
\renewcommand{\thesection}{\Alph{section}}
\renewcommand\thefigure{\thesection.\arabic{figure}} 
\setcounter{figure}{0}
\renewcommand\thetable{\thesection.\arabic{table}} 
\setcounter{table}{0}

This section details the license agreements of the scientific artifacts used in this work. The dataset COSMOSQA \citep{Huang2019CosmosQM} has BSD 3-Clause License. The datasets RACE++ \citep{Lai2017RACELR} and ReClor \citep{Yu2020ReClorAR}  are freely available with the limitation on the latter that it can only be used for non-commercial research purposes. Huggingface transformer models are released under: Apache License 2.0. All the scientific aritfacts are consistent with their intended uses.

\newpage
\renewcommand{\thesection}{Appendix \Alph{section}}
\section{Low Entropy Examples} \label{sec:more_ex}
\renewcommand{\thesection}{\Alph{section}}
\renewcommand\thefigure{\thesection.\arabic{figure}} 
\setcounter{figure}{0}
\renewcommand\thetable{\thesection.\arabic{table}} 
\setcounter{table}{0}

\begin{figure}[h!]
    \centering
    \includegraphics[width=1.0\columnwidth]{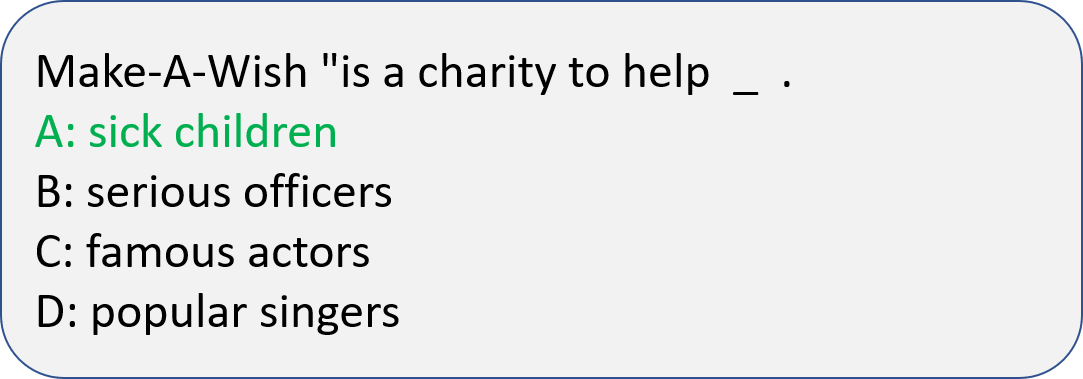}
\end{figure}

\begin{figure}[h!]
    \centering
    \includegraphics[width=1.0\columnwidth]{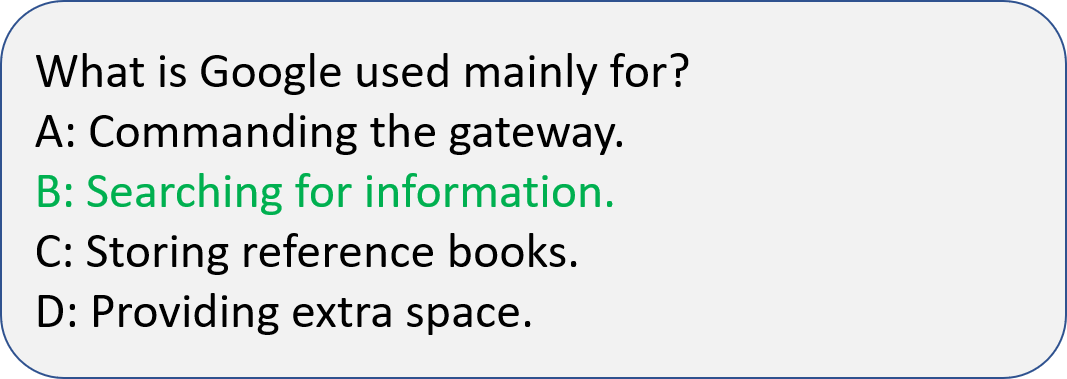}
\end{figure}

\begin{figure}[h!]
    \centering
    \includegraphics[width=1.0\columnwidth]{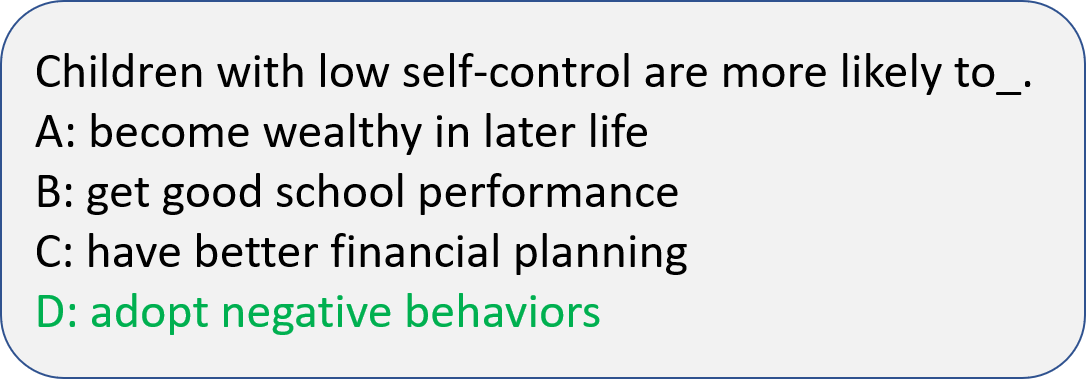}
\end{figure}

\begin{figure}[h!]
    \centering
    \includegraphics[width=1.0\columnwidth]{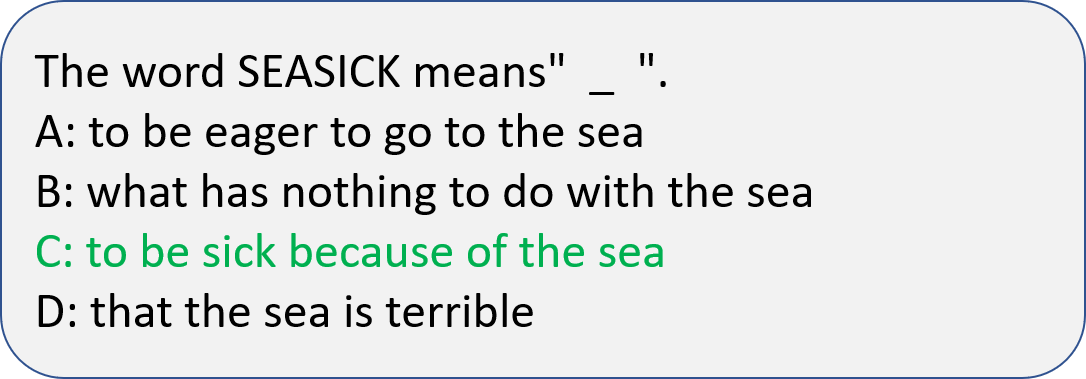}
\end{figure}

\begin{figure}[h!]
    \centering
    \includegraphics[width=1.0\columnwidth]{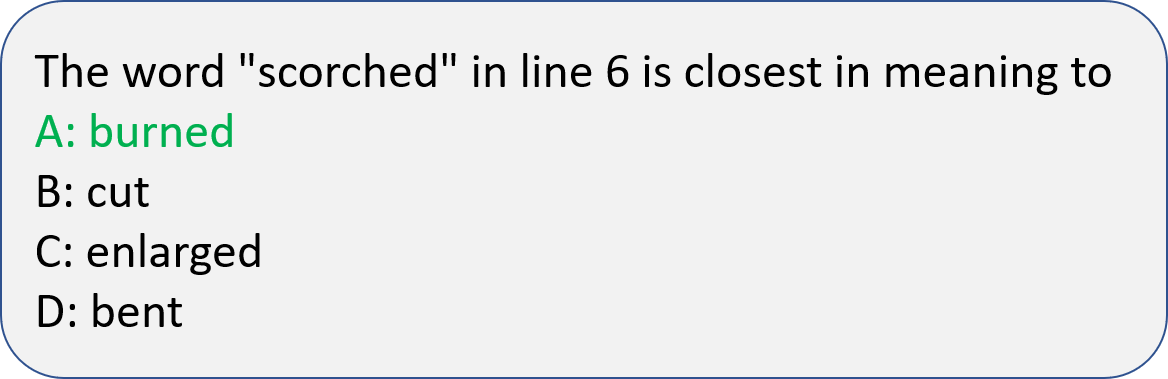}
\end{figure}

\end{document}